\documentclass[runningheads]{llncs}
\usepackage{graphicx}
\usepackage{amsfonts}
\usepackage{amsmath}
\usepackage{multirow}
\usepackage{booktabs}
\usepackage{mathtools}
\usepackage{caption}
\usepackage{subcaption}
\begin{document}
\title{If You Like It, GAN It\protect\\ Probabilistic Multivariate Times Series Forecast With GAN}
\titlerunning{If You Like It, GAN It}
%
\author{Alireza Koochali\inst{1,2,3}\orcidID{0000-0001-7370-9369} \and
Andreas Dengel\inst{2,3}\orcidID{0000-0002-6100-8255} \and
Sheraz Ahmed\inst{2}\orcidID{0000-0002-4239-6520}}
\authorrunning{A. Koochali et al.}
%
\institute{IAV GmbH, Carnotstr. 1, 10587 Berlin, Germany\\ 
\email{alireza.koochali@iav.de} \and
 DFKI GmbH, Trippstadter Str. 122, 67663 Kaiserslautern, Germany\\
\email{\{alireza.koochali,andreas.dengel,sheraz.ahmed}@dfki.de\} \and
University of Kaiserslautern, Erwin-Schrödinger-str. 52, 67663 Kaiserslautern, Germany\\
\email{akoochal@rhrk.uni-kl.de}}
\maketitle              
\begin{abstract}
The contribution of this paper is two-fold. First, we present ProbCast - a novel probabilistic model for multivariate time-series forecasting. We employ a conditional GAN framework to train our model with adversarial training. Second, we propose a framework that lets us transform a deterministic model into a probabilistic one with improved performance. The motivation of the framework is to either transform existing highly accurate point forecast models to their probabilistic counterparts or to train GANs stably by selecting the architecture of GAN's component carefully and efficiently. We conduct experiments over two publicly available datasets namely electricity consumption dataset and exchange-rate dataset. The results of the experiments demonstrate the remarkable performance of our model as well as the successful application of our proposed framework.

\keywords{Time-series, Generative Adversarial Networks, Forecasting, Probabilistic, Prediction}
\end{abstract}
\section{Introduction}
A large sector of industry such as health care, automotive industry, aerospace industry, and weather forecast deals with time-series data in their operations. The knowledge about what will happen in the future is essential to make genuine decision and accurate forecasting of future values is a key to the success. Hence, a huge body of research is dedicated to address the forecasting problem. An overview of various researches on forecasting problem is provided in ~\cite{Mahalakshmi2016survey}. Currently, the field is dominated by point prediction methods which are easy to understand. However, these deterministic models report the mean of possible outcomes and cannot reflect the inherent uncertainty exists in the real world. The probabilistic forecast models are devised to answer these shortcomings. These models try to quantify the uncertainty of the predictions by forming a  probability distribution over possible outcomes~\cite{gneiting2014probabilistic}.

In this paper, we propose ProbCast, a new probabilistic forecast model for multivariate time-series based on Conditional Generative Adversarial Networks (GANs). Conditional GANs are a class of NN-based generative models that enables us to learn conditional probability distribution given a dataset. ProbCast is trained using a Conditional GAN setup to learn the probability distribution of future values conditioned on the historical information of the signal.

While GANs are powerful methods for learning complex probability distributions, they are notoriously hard to train. The training process is very unstable and quite dependant on careful selection of the model architecture and hyperparameters~\cite{Goodfellow-et-al-2016}. In addition to ProbCast, we suggest a framework to transform an existing deterministic forecaster - which is comparatively easy to train - into a probabilistic one which excels its predecessor. By using the proposed framework, the space for searching GAN's architecture becomes considerably smaller. Thus, this framework provides an easy way to adapt highly accurate deterministic models to construct useful probabilistic models without compromising the accuracy by exploiting the potential of GANs.

In summary, the main contributions of this article are as follows:
\begin{itemize}
    \item We introduce ProbCast, a novel probabilistic model for multivariate time-series forecasting. Our method employs conditional GAN setup to train a probabilistic forecaster.
    \item We suggest a framework to transform a point forecast model into a probabilistic model. This framework eases the process of replacing the deterministic model with probabilistic ones.
    \item We conduct experiments on two publicly available datasets and report the results which shows the superiority of the ProbCast. Furthermore, we demonstrate that our framework is capable of transforming a point forecast method into a probabilistic model with improved accuracy.
\end{itemize}

\section{Related Work}

Due to the lack of a standard evaluation method for GANs, initially, they were applied to domains in which their results are intuitively assessable e.g. images. However, recently they have been applied to time-series data. Currently GANs are applied to various domains for generating realistic time-series data including health care~\cite{esteban2017real,golany2019pgans,haradal2018biosignal,nikolaidis2019augmenting,ye2019ecg} ,finance~\cite{wiese2019deep,wiese2020quant} , and energy industry~\cite{chen2018model,fekri2020generating,zhanggenerative}. In~\cite{yoon2019time}, authors combine GAN and auto-regressive models to improve sequential data generation. Ramponi et al.~\cite{ramponi2018t} condition a GAN on timestamp information to handle irregularly sampling.

Furthermore, researchers have used conditional GAN to build probabilistic forecasting models. Koochali et al.~\cite{koochali2019forgan} use Conditional GAN to build a probabilistic model for univariate time-series. They use Long Short-Term Memory (LSTM) in GAN's component and test their method on a synthetic dataset as well as two publicly available datasets. In~\cite{zhang2019stock} authors utilize LSTM and Multi-Layer Perceptron (MLP) in a conditional GAN structure to forecast the daily closing price of stocks. The authors combine the Mean Square Error (MSE) with the generator loss of a GAN to improve performance. Zhou et al.~\cite{zhou2018stock}, employ LSTM and convolutional neural network (CNN) in an adversarial training setup to forecast the high-frequency stock market. To guarantee satisfying predictions, this method minimizes the forecast error in the form of Mean absolute error (MAE) or MSE during training in conjunction with the GAN value function. Lin et al.~\cite{lin2018pattern} propose a pattern sensitive forecasting model for traffic flow which can provide accurate predictions in unusual states without compromising its performance in usual states. This method uses conditional GAN with MLP in its structure and adds two error terms to standard generator loss. The first term specifies forecast error and the second term expresses reconstruction error. Kabir et al.~\cite{kabir2019partial} make use of adversarial training for quantifying the uncertainty of the electricity price with prediction interval. This line of research is more aligned with the method we presented in this article however the methods suggested in~\cite{lin2018pattern,zhang2019stock,zhou2018stock} include a point-wise loss function into the GAN loss function. Minimizing suggested loss functions would decrease statistical error values, such as RMSE, MAPE, and MSE. However, they encourage the model to learn the mean of possible outcomes instead of the probability distribution of future value. Hence, their probabilistic forecast can be misleading despite the small statistical error.

\section{Background}
Here, we work with a multivariate time-series $X = \{X_{0},X_{1},...,X_{T}\}$ where each $X_{t} = \{x_{t,1},x_{t,2},...,x_{t,f}\}$ is a vector with size $f$ equal to the number features. In this paper $x_{t,f}$ refers to data point at time step $t$ of feature $f$ and $X_{t}$ points to feature vector at time step $t$. The goal is to model $P(X_{t+1}|X_{t},..,X_{0})$, the probability distribution for $X_{t+1}$ given historical information $\{X_{t},..,X_{0}\}$. 

\subsection{Mean regression forecaster}

To address the problem of forecasting, we can take the predictive view of regression~\cite{gneiting2014probabilistic}. Ultimately, the regression analysis aims to learn the conditional distribution of a response given a set of explanatory variables~\cite{hothorn2014conditional}. The mean regression methods are deterministic methods which are concerned with accurately predicting the mean of possible outcome i.e. $\mu(P(X_{t+1}|X_{t},..,X_{0}))$. There is a broad range of mean regression methods available in the literature however, all of them are unable to reflect uncertainty in their forecasts. Hence, their results can be unreliable and misleading in some cases.

\subsection{Generative Adversarial Network}

In 2014, Goodfellow et al.~\cite{goodfellow2014generative} introduce a powerful generative model called Generative Adversarial Network (GAN).
 GAN can implicitly learn probability distribution which describes a given dataset i.e. $P(data)$ with high precision. Hence, it is capable of generating artificial samples with high fidelity. The GAN architecture is composed of two neural networks namely generator and discriminator. These components are trained simultaneously in an adversarial process. In the training process, first, a noise vector $z$ is sampled from a known probability distribution $P_{noise}(z)$ and fed into generator. Then, generator transforms $z$ from $P_{noise}(z)$ to a sample which follows $P_{data}$. On the other hand, discriminator checks how well generator is performing by comparing generator's outputs with real samples from the dataset. During training, this two-player minimax game is set in motion by optimizing the following value function:
\begin{equation}
\begin{split}
\min_{G} \max_{D} V(D,G) ={} & \mathbb{E}_{x\sim P_{data}(x)}[log(D(x))] +{} \\
& \mathbb{E}_{z\sim P_{noise}(z)}[log(1-D(G(z)))].
\end{split}
\end{equation}

However, GAN's remarkable performance does not acquire easily. Their training process is quite unstable and careful selection of GAN's architecture and hyperparameters is vital for stabilizing training process~\cite{Goodfellow-et-al-2016}. Since we should search for the optimal architecture of generator and discriminator simultaneously, it is normally a cumbersome and time-consuming task to find a perfect combination of structures in a big search space.

\subsection{Conditional GAN}

Conditional GAN~\cite{mirza2014conditional} enables us to incorporate auxiliary information called condition into the process of data generation. In this method, we provide an extra piece of information like labels to both generator and discriminator. The generator must respect the condition while synthesizing a new sample because the discriminator considers the given condition while it checks the authenticity of its input. The new value function $V (G,D)$ for this setting is:
\begin{equation}
\begin{split}
\min_{G} \max_{D} V(D,G) ={} & \mathbb{E}_{x\sim P_{data}(x)}[log(D(x|y))] +{} \\
& \mathbb{E}_{z\sim P_{noise}(z)}[log(1-D(G(z|y)))].
\end{split}
\end{equation}
After training a Conditional GAN, the generator learns implicit the probability distribution of data given condition i.e. $P(data|condition)$


\begin{figure}
 \includegraphics[width=\textwidth]{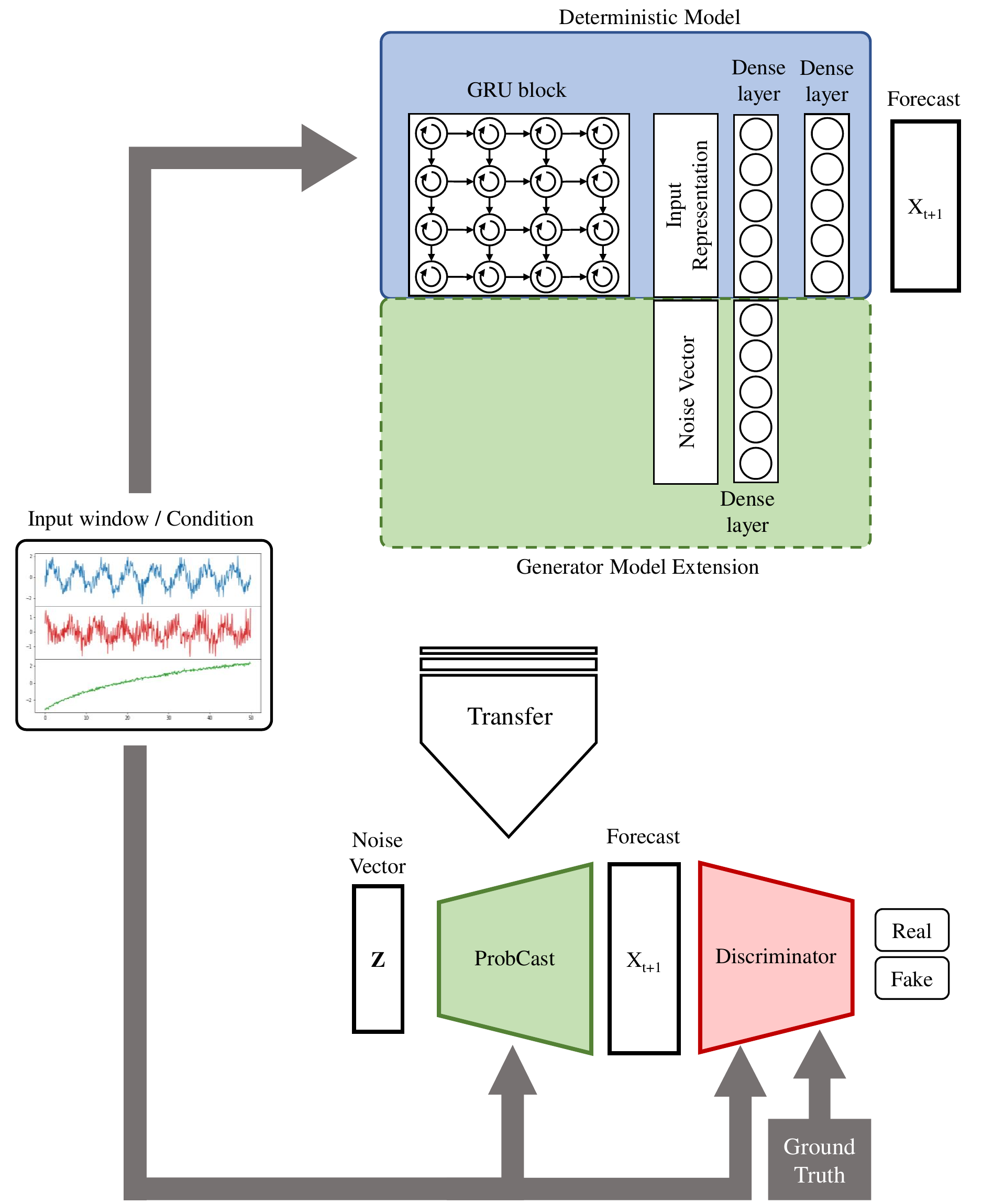}
 \caption{The demonstration of the proposed framework and adversarial training setup. The pipeline is followed from top to bottom. First, we search for the optimal architecture of the deterministic model. The deterministic model consists of a GRU block for learning input window representation and two dense layers to map the representation to the forecast. Then, the noise vector $z$ is integrated into the deterministic model to build the generator. Finally, the generator is trained using suitable discriminator in a conditional GAN setup to obtain ProbCast. } \label{fig:big}
 \end{figure}

\section{Methodology}
\subsection{ProbCast: The proposed multivariate forecasting model }

In this article, we consider Conditional GAN as a method for training a probabilistic forecast model using adversarial training. In this perspective, the generator is our probabilistic model (i.e. ProbCast) and the discriminator provides required gradient for optimizing ProbCast during training. To learn $P(X_{t+1}|X_{t},..,X_{0})$, the historical information $\{X_{t},..,X_{0}\}$ is used as the condition of our Conditional GAN and the generator is trained to generate $X_{t+1}$. Hence, the probability distribution which is learned by generator corresponds to $P(X_{t+1}|X_{t},..,X_{0})$ i.e. our target distribution. The value function which we used for training the ProbCast (indicated as PC) is:

\begin{equation}
\begin{split}
\min_{PC} \max_{D} V(D,PC) ={} & \mathbb{E}_{X_{t+1}\sim P_\text{data}(X_{t+1})}[\text{ log}\, D(X_{t+1}|X_{t},..,X_{0})] +{} \\
& \mathbb{E}_{z\sim P_{z}(z)}[\text{ log}\,(1-D(PC(z|X_{t},..,X_{0})))].
\end{split}
\end{equation}

\subsection{The proposed framework for converting deterministic model to probabilistic model}
By stepping in the realm of multivariate time-series, other challenges need to be addressed too. In the multivariate setting, we require more complicated architecture to figure out dependencies between features and forecast future with high accuracy. Furthermore, as previously mentioned, GANs require precise hyperparameter tuning to have a stable training process. Considering required network complexity for handling multivariate time-series data, it is very cumbersome or in some cases impossible to find suitable generator and discriminator architecture concurrently which performs accurately. To address this problem, we propose a new framework for building a probabilistic forecaster based on a deterministic forecaster using GAN architecture.

In this framework, we build the generator based on the architecture and hyper-parameters of the deterministic forecaster and train it using appropriate discriminator architecture. In this fashion, we can perform the task of finding an appropriate generator and discriminator architecture separately which results in simplification of the GAN architecture search process. In other words, by using this framework, we can transform an existing accurate deterministic model into a probabilistic model with increased precision and better alignment with the real world.

\subsection{Train pipeline}
Figure~\ref{fig:big} demonstrates the proposed framework as well as conditional GAN setup for training ProbCast. First, we build an accurate point forecast model by searching for the optimal architecture of the deterministic model. In the case that a precise point forecast model exist, we can skip the first step and use existing model. Then, we need to integrate the noise vector $z$ into the deterministic model architecture. In our experiments, we get the best results when we insert the noise vector into the later layers of the network, letting earlier layers of network learn the representation of the input window. Finally we train this model using adversarial training to acquire our probabilistic forecast model i.e. ProbCast.

With generator architecture at hand, we only need to search for an appropriate time-series classifier to serve as the discriminator during the training of GAN. By reducing the search space of GAN architecture to discriminator only, we can find a discriminator structure efficiently which is capable of training the ProbCast with superior performance in comparison to the deterministic model. The following steps summarize the framework:
\begin{enumerate}
    \item Employ an accurate deterministic model:
    \begin{enumerate}
        \item Either use an existing model
        \item Or search for an optimal deterministic forecaster
    \end{enumerate}
    \item Structure the generator based on deterministic model architecture and incorporate noise vector into the network, preferably into later layers.
    \item Search for an optimal discriminator structure and train the generator using the that.
\end{enumerate}


    
    

\section{Experiment}
\subsection{Datasets}

We tested our method on two publicly available datasets namely electricity and exchange-rate\footnote{We used dataset from https://github.com/laiguokun/multivariate-time-series-data as they were prepared by the authors of~\cite{lai2018modeling}}. \textbf{Electricity dataset} consists of electricity consumption of 321 clients in KWh which is collected every 15 minutes between 2012 and 2014. The dataset is converted to reflect hourly consumption. \textbf{Exchange-rate dataset} contains daily exchange-rate of eight countries namely Australia, British, Canada, Switzerland, China, Japan, New Zealand, and Singapore which is collected between 1990 to 2016. Table~\ref{tab:dataset} lists the properties of these two datasets.
For our experiments, we used $75\%$ of the datasets for training, $5\%$ for validation and $20\%$ for testing.

\begin{table}
    \centering
    \begin{tabular}{lcc}
        \toprule
                    & ~~~\textbf{Electricity dataset}~~~      &~~~\textbf{Exchange-rate dataset}~~~\\
        \midrule
        Dataset length          & 62,304 & 7,588\\
        Number of feature        & 321    & 8 \\
        Sample rate              & 1 hour & 1 day\\
        \bottomrule
         & & \\
    \end{tabular}
    \caption{The properties of datasets}
    \label{tab:dataset}
\end{table}

\subsection{Setup}
\begin{figure}
\includegraphics[width=\textwidth]{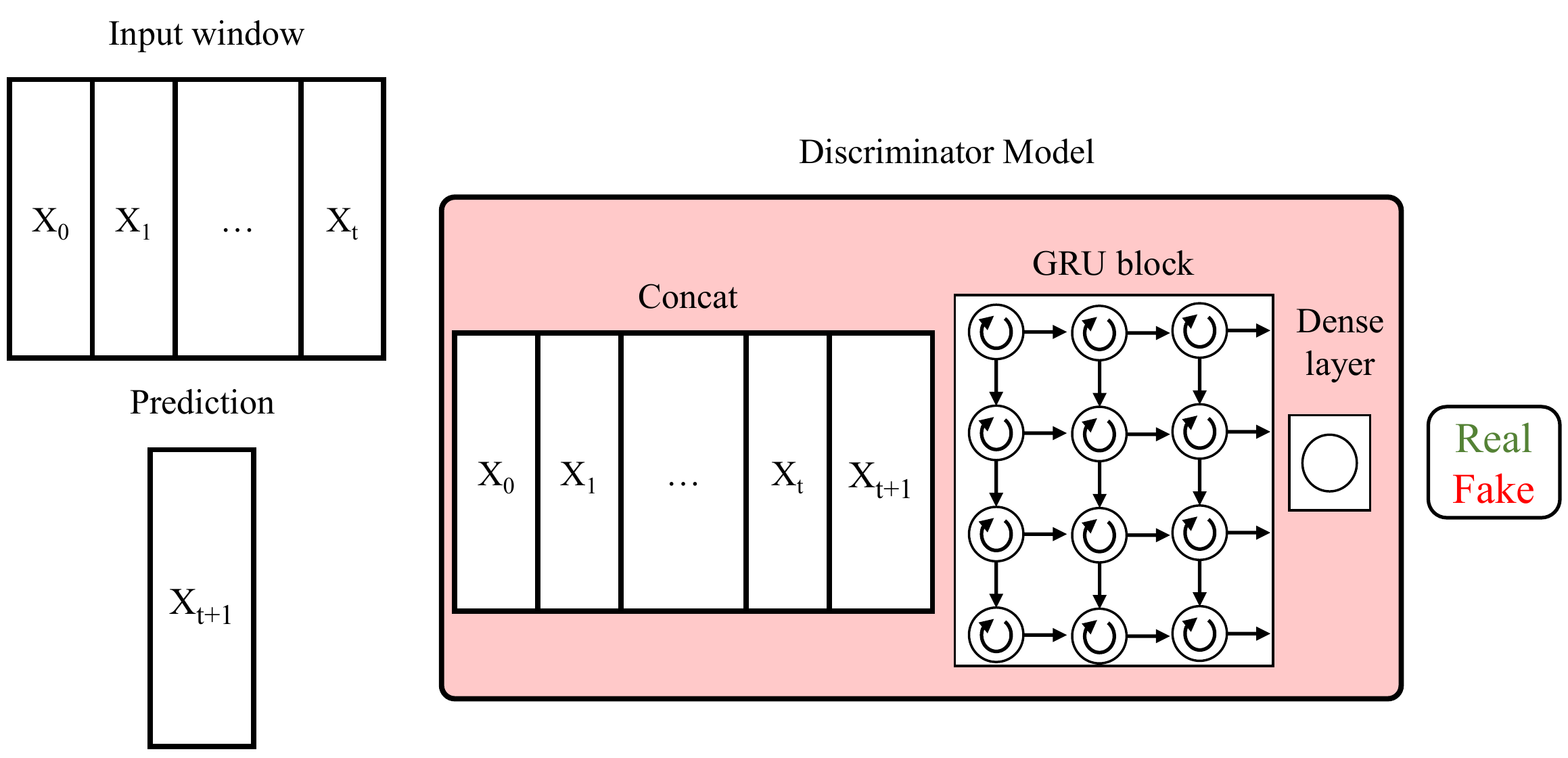}
\caption{The discriminator architecture of our conditional GAN. The number of layers and cells in GRU block are hyperparameters.} \label{fig:dis}
\end{figure}

In each of our experiments, first, we run architecture search to find an accurate deterministic model. For training the deterministic model, we employed MAE as a loss function. In figure~\ref{fig:big}, the architecture of deterministic model is indicated. We used a Gated recurrent unit (GRU)~\cite{cho2014learning} block to learn the representation of the input window. Then, the representation passes through two dense layers to map from representation to forecast. We adopt the architecture of the most precise deterministic model which we found to build the ProbCast by concatenating noise vector to GRU block output (i.e. representation) and extending MLP block as shown in Figure~\ref{fig:big}. Finally, we search for the optimal architecture of the discriminator (Figure~\ref{fig:dis}) and train the ProbCast. The discriminator concatenates $X_{t+1}$ to the end of input window and constructs $\{X_{t+1},X_{t},..,X_{0}\}$. Then it utilizes a GRU block followed by two layers of MLP to inspect the consistency of this window. We use the genetic algorithm to search for the optimal architecture. We code our method using Pytorch~\cite{paszke2017automatic}.

\begin{table}
    \centering
    \begin{tabular}{lcc}
        \toprule
        \textbf{Generator}~~&  &\\
        \textbf{Hyperparameters}~~& ~~\textbf{Electricity}~~ &~~\textbf{Exchange-rate}\\
        \midrule
         Input windows size                  & 174                       & 170\\
         Noise size                          & 303                       & 183\\
         Number of GRU layers                & 1                       & 1\\
         Number of GRU cells in each layer    & 119                       & 119\\
        \toprule
         \textbf{Discriminator}~~&  &\\
         \textbf{Hyperparameters}~~&  &\\
        \midrule
         Number of GRU layers                & 3                       & 1\\
         Number of GRU cells in each layer    & 146                       & 149\\
        \bottomrule
         & & \\
    \end{tabular}
    \caption{The List of the hyperparameters alongside their optimal values for each experiments.}
    \label{tab:hyper}
\end{table}

\subsection{Evaluation Metric}
To report the performance of the ProbCast, we used the negative form of Continues Ranked Probability Score~\cite{gneiting2007strictly} (denoted by $CRPS^*$) as the metric. The $CRPS^*$ reflects the sharpness and calibration of a probabilistic method. It is defined as follow:
\begin{equation}
    CRPS^*(F,x) = E_{F}|X-x| - \frac{1}{2}E_{F}|X-X'|,
    \label{eq:crps}
\end{equation}

where X and X' are independent copies of a random variable from probabilistic forecaster F and x is the ground truth. The $CRPS^*$ provides a direct way to compare deterministic and probabilistic models. In the case of the deterministic forecaster, the $CRPS^*$ reduces to Mean Absolute Error (MAE) which is a commonly used point-wise error metric.
In other words, in a deterministic setting, the $CRPS^*$ is equivalent to MAE:
\begin{equation}
    MAE(x,\hat{x}) = E|\hat{x}-x|,
\end{equation}
where x is the ground truth and $\hat{x}$ is the point forecast. After the GAN training concluded, we calculate the $CRPS^*$ of the ProbCast and deterministic model. To calculate $CRPS^*$ for ProbCast using equation~\ref{eq:crps}, we sample it 200 times (100 times for each random variable). 

\section{Results and Discussion}
Table~\ref{tab:hyper} presents optimal hyperparameters we found for each dataset using our framework during the experiments and table~\ref{tab:res} summarizes our experiments' results presenting $CRPS^*$ of the best deterministic model and the ProbCast for each dataset.

In the experiment with the electricity dataset, the ProbCast is more accurate than the deterministic model while it has an almost identical structure. Furthermore, this experiment shows that our model can provide precise forecasts for multivariate time-series even when the number of features is substantial. In the exchange-rate experiment, the ProbCast outperforms its deterministic predecessor again despite structural similarities. We can also observe that our method works well even though the dataset is considerably smaller in comparison to the previous experiment.

Furthermore, it confirms that our framework is capable of transforming a deterministic model to a probabilistic model which is more accurate than its predecessor. The question now arises: Considering the sensitiveness of GAN to the architecture of its components, why employing deterministic model architecture to define the ProbCast works fine while it is borrowed from a totally different setup? We think that the deterministic model provides us an architecture that is capable of learning good representation from the input time window. Since the model is trained to learn the mean of possible outcomes, these representations contain a distinctive indicator of where the target distribution is located. With the help of these indicators, the MLP block learns to accurately transform the noise vector $z$ to the probability distribution of future values.

\begin{table}[t]
    \centering
    \begin{tabular}{lcc}
        \toprule
        \textbf{Dataset}                       & ~~~\textbf{Deterministic Model}~~~      &~~~\textbf{ProbCast}~~~\\
        \midrule
        Electricity                          & 235.96                       & 232.00\\
        Exchange-rate                          & $1.04\times10^{-2}$                       & $8.66\times10^{-3}$\\
        \bottomrule
         & & \\
    \end{tabular}
    \caption{The results of the experiments for the deterministic model and the ProbCast reported in $CRPS^*$.}
    \label{tab:res}
\end{table}

\section{Conclusion and Future works}
In this paper, we present ProbCast, a probabilistic model for forecasting one step ahead of multivariate time-series. We employ the potential of conditional GAN in learning conditional probability distribution to model probability distribution of future values given past values i.e. $P(X_{t+1}|X_{t},..,X_{0})$.

Furthermore, we propose a framework to efficiently find the optimal architecture of GAN's components. This framework builds the probabilistic model upon a deterministic model to improve its performance. Hence, it enables us to search for optimal architecture of generator and discriminator separately. Furthermore, it can transform an existing deterministic model into a probabilistic model with increased precision and better alignment with the real world.

We assess the performance of our method on two publicly available datasets. The exchange-rate dataset is a small dataset with a few numbers of features while Electricity dataset the is bigger with a considerably larger number of features. We compare the performance of the ProbCast with its deterministic equivalent. In both experiments, our method outperforms its counterpart. The results of the experiments demonstrate that the ProbCast can learn patterns precisely from a small set of data and at the same time, it is capable of figuring out the dependencies between a large number of features and forecast future values accurately in the presence of a big dataset. Furthermore, the results of the experiments indicate the successful application of our framework which paves the way for a systematic and straightforward approach to exchange currently used deterministic models with a probabilistic model to improve accuracy and obtain realistic forecasts.

The promising results of our experiments signify a great potential in probabilistic forecasting using GANs and suggest many new frontiers to further push the research in this direction. For instance, we employ vanilla GAN for our research while there are a lot of modifications suggested for improving GANs in recent years. One possible direction is to apply these modifications and inspect the improvement in the performance of the ProbCast. The other direction is experimenting with more sophisticated architectures for generator and discriminator. Finally, we only use the knowledge from the deterministic model to shape the generator. It would be interesting to push this direction and try to incorporate more knowledge from the deterministic model into the GAN training process to improve and optimize the probabilistic model.

\bibliographystyle{splncs04}
\bibliography{lite.bib}

\end{document}